\documentclass[conference]{IEEEtran}
\IEEEoverridecommandlockouts
\usepackage{cite}
\usepackage{amsmath,amssymb,amsfonts}
\usepackage{graphicx}
\usepackage{textcomp}
\usepackage{xcolor}
\usepackage[hidelinks]{hyperref}
\usepackage{multirow}
\usepackage{booktabs}
\usepackage{threeparttable}
\usepackage{adjustbox}
\usepackage{diagbox}
\usepackage{subcaption}
\usepackage{tabularx}
\usepackage{algorithmicx}
\usepackage[ruled,linesnumbered]{algorithm2e}
\newtheorem{theorem}{{Theorem}}
\newtheorem{coro}{{Corollary}} 
\newenvironment{pf}
{\par\indent{\it Proof.} }
{\hfill$\blacksquare$\par}

\begin{document}

\title{Lite-RVFL: A Lightweight Random Vector Functional-Link Neural Network for Learning Under Concept Drift}

\author{\IEEEauthorblockN{Songqiao Hu}
\IEEEauthorblockA{\textit{Department of Automation} \\
\textit{Tsinghua University}\\
Beijing, China \\
hsq23@mails.tsinghua.edu.cn}
\and
\IEEEauthorblockN{Zeyi Liu}
\IEEEauthorblockA{\textit{Department of Automation} \\
\textit{Tsinghua University}\\
Beijing, China \\
liuzy21@mails.tsinghua.edu.cn}
\and
\IEEEauthorblockN{Xiao He}
\IEEEauthorblockA{\textit{Department of Automation} \\
\textit{Tsinghua University}\\
Beijing, China \\
hexiao@tsinghua.edu.cn
}
\thanks{This work was supported by National Natural Science Foundation of China under grants 62473223 and 624B2087,  and Beijing Natural Science Foundation under grant L241016 (Corresponding author: Xiao He).}
}

\maketitle

\begin{abstract}
The change in data distribution over time, also known as concept drift, poses a significant challenge to the reliability of online learning methods. Existing methods typically require model retraining or drift detection, both of which demand high computational costs and are often unsuitable for real-time applications. To address these limitations, a lightweight, fast and efficient random vector functional-link  network termed Lite-RVFL is proposed, capable of adapting to concept drift without drift detection and retraining. Lite-RVFL introduces a novel objective function that assigns weights exponentially increasing to new samples, thereby emphasizing recent data and enabling timely adaptation. Theoretical analysis confirms the feasibility of this objective function for drift adaptation, and an efficient incremental update rule is derived. Experimental results on a real-world safety assessment task validate the efficiency, effectiveness in adapting to drift, and potential to capture temporal patterns of Lite-RVFL. The source code is available at \url{https://github.com/songqiaohu/Lite-RVFL}.
\end{abstract}


\section{Introduction}

Online learning holds significant potential for handling continuous data streams and has been widely adopted across diverse application domains \cite{hoi2021online}. Its primary objective is to continuously update models using streaming data samples to enhance prediction performance. However, changes in data distribution over time pose considerable challenges to the effectiveness of online learning methods. This issue, known as concept drift, can be classified into virtual drift, where the feature distribution changes, and real drift, where the decision boundary shifts \cite{lu2018learning}. If not properly addressed, concept drift can result in degraded prediction accuracy. A notable example is real-time safety assessment (RTSA) of dynamic systems, which typically involves building models to predict and evaluate system safety based on data collected in real time from system sensors \cite{liu2022online, liu2023real}. As systems continue to operate, the statistical properties of the collected sensor data may change due to factors such as component aging, variations in operating conditions, or environmental fluctuations \cite{vzliobaite2016overview, li2025dynamic}, leading to degraded model performance. Therefore, how to effectively address concept drift holds substantial research value and practical importance.


Recent studies have focused on handling concept drift, which can be divided into active and passive approaches \cite{han2022survey}. Active methods detect significant changes in data feature distribution or model performance using statistical analysis. If drift is detected, the model is typically retrained using the most recent samples to quickly adapt to the new concept. Typical drift detectors include ADWIN \cite{bifet2007learning}, HDDM \cite{frias2014online}, and CADM+ \cite{hu2024cadm}. In contrast, passive methods, commonly found in ensemble learning, naturally adapt to drift by continuously updating, replacing, or re-weighting ensemble members, without explicitly tracking when drift occurs. For example, in references \cite{zhang2020reinforcement, jiao2022dynamic, lu2019adaptive}, a new classifier is trained using a fixed number of samples at regular intervals, replacing either the classifier with the worst performance or the oldest classifier in the ensemble.


However, both active and passive approaches typically require retraining the classifier, which demand significantly higher computational costs compared to incremental updates. In active methods, additional overhead is also introduced by the drift detection mechanisms. Such computational demands make these approaches difficult to deploy in real-time or resource-constrained scenarios.
In addition, in our previous work \cite{hu2025performance}, we established a theoretical lower bound on the expected accuracy of an ensemble classifier relative to its base classifiers. However, due to the asynchronous and unknown timing of retraining triggered by drift detection in individual base classifiers, it becomes challenging to derive a theoretical lower bound for the ensemble directly with respect to the data stream.
Therefore, it is of considerable interest to explore a framework that can adapt to evolving data distributions through incremental updates, without relying on drift detection and retraining.


To address the aforementioned challenges, this paper proposes {Lite-RVFL}, a lightweight, fast and efficient random vector functional-link  network (RVFL) designed for concept drift adaptation, based on the original RVFL \cite{pao1994learning}. To achieve this, a novel objective function is proposed that assigns exponentially increasing weights to new samples, thereby emphasizing more recent data. Furthermore, the incremental update expression for Lite-RVFL is derived, facilitating its efficient adaptation to non-stationary environments. The main contributions of this work are summarized as follows:

\begin{itemize}
    \item[(1)] A lightweight RVFL for concept drift adaptation termed Lite-RVFL is proposed, in which a novel objective function is introduced based on the original RVFL. This formulation assigns exponentially increasing weights to new samples, which facilitates the capture of temporal patterns in sequential data.
    \item[(2)] 
    The incremental update rule for Lite-RVFL is established, along with a theoretical framework that characterizes its emphasis on new samples. Theoretical analysis shows that Lite-RVFL maintains a nearly constant level of attention to the most recent samples.

    \item[(3)] Comprehensive experiments are conducted on a RTSA task.  The results demonstrate that Lite-RVFL achieves performance comparable to RVFL integrated with a drift detector, while maintaining nearly the same computational efficiency to that of the original RVFL 
\end{itemize}

The rest of this article is organized as follows. In Section \ref{sec:methods}, the technical and
 theoretical details of Lite-RVFL are presented. In Section \ref{sec:experiments},
 the experimental results and analysis are shown, and Section \ref{sec:conclusion}
 provides the conclusion and further work.

\section{The Proposed Lite-RVFL}
In this section, the theoretical details of Lite-RVFL are presented, and a theoretical comparison is made with a similar approach.
\label{sec:methods}
\subsection{Design of Lite-RVFL}
Assume that we have collected a lot of samples $\{{\boldsymbol x_i}, s_i\}$ in the offline stage, where $s_i\in \{1, 2, \cdots, m\}$, for $i\in \{1, 2, \cdots, N\}$. Let $\tilde{\boldsymbol{x}} \in \mathbb{R}^{\left(m+\left(N_1 N_2\right)\right) \times 1}$ represent a combination of $\boldsymbol{x}$ and its enhancement features. Additionally, let \( \boldsymbol{\tilde{s}} \) denote the one-hot vector representation of \( s \):
\begin{equation}
\label{eq:W_e_b_e}
\tilde{\boldsymbol{x}}=\left[\begin{array}{c}
\boldsymbol{x} \\
\boldsymbol{Z}
\end{array}\right]=\left[\begin{array}{c}
\boldsymbol{x} \\
\phi\left(\boldsymbol{x}^\top \boldsymbol{W_{e_1}
}+\boldsymbol{b_{e_1}}\right)\\
\phi\left(\boldsymbol{x}^\top \boldsymbol{W_{e_2}}+\boldsymbol{b_{e_2}}\right)\\
\vdots\\
\phi\left(\boldsymbol{x}^\top \boldsymbol{W_{e_{N_1}}}+\boldsymbol{b_{e_{N_1}}}\right)
\end{array}\right], 
\end{equation}
\begin{equation}
\boldsymbol{\tilde{s}} =
\begin{cases}
[1, 0, 0, \dots, 0], & \text{if } s = 1 \\
[0, 1, 0, \dots, 0], & \text{if } s = 2 \\
\vdots & \\
[0, 0, \dots, 0, 1], & \text{if } s = m \\
\end{cases},
\end{equation}
where $\phi(\cdot): \mathbb{R} \to \mathbb{R}$ is an activation function, both \(\boldsymbol{W}_e=\left[\boldsymbol{W}_{e_1}, \boldsymbol{W}_{e_2}, \ldots, \boldsymbol{W}_{e_{N_1}}\right]\) and \(\boldsymbol{b}_e=\left[\boldsymbol{b}_{e_1}, \boldsymbol{b}_{e_2}, \ldots, \boldsymbol{b}_{e_{N_1}}\right]\) are randomly initialized and subsequently fixed \cite{malik2023random, zhang2016comprehensive}. If the input of $\phi(\cdot)$ is a matrix, the function $\phi(\cdot)$ is applied element-wise to the matrix. Let the extended data matrix be denoted as  
\begin{equation}
\label{eq:A_S}
\boldsymbol{{A}}=[\tilde{\boldsymbol{x}}_1, \tilde{\boldsymbol{x}}_2, \cdots, \tilde{\boldsymbol{x}}_N]^\top, \boldsymbol{{S}}=[\boldsymbol{\tilde{s}}_1^\top, \boldsymbol{\tilde{s}}_2^\top, \cdots, \boldsymbol{\tilde{s}}_N^\top]^\top.
\end{equation}

Let the weights of the time-series samples increase, with the weight of each subsequent sample being \( \theta \) times the weight of the previous sample, where \( \theta > 1 \). This leads the classifier to place greater emphasis on the most recent concepts. The training of the classifier can then be summarized as solving the optimization problem in Eq.~(\ref{eq:optimization}).
\begin{equation}
\label{eq:optimization}
\boldsymbol{W_b}=\underset{\boldsymbol{W}}{\arg \min } \quad \lambda\left\|\boldsymbol{W}\right\|_2^2+\left\|\boldsymbol{T}_N\left(\boldsymbol{{A}}\boldsymbol{W}-\boldsymbol{{S}}\right)\right\|_2^2,
\end{equation}
where \(\boldsymbol{T}\) is the weights matrix of samples:
\begin{equation}
\boldsymbol{T}_N = 
\begin{bmatrix}
1 & 0 & \cdots & 0 \\
0 & \theta & \cdots & 0 \\
\vdots & \vdots & \ddots & \vdots \\
0 & 0 & \cdots & \theta^{N-1} \\
\end{bmatrix}.
\end{equation}
$\boldsymbol{W}_b$ can then be obtained as follows:
\begin{equation}
\label{eq:W_b}
\boldsymbol{W_b}=\left(\lambda \boldsymbol{I}+\boldsymbol{{A}}^\top\boldsymbol{T}_N^\top\boldsymbol{T}_N \boldsymbol{{A}}\right)^{-1} \boldsymbol{{A}}^\top\boldsymbol{T}_N^\top\boldsymbol{T}_N \boldsymbol{{S}}.
\end{equation}
Based on $\boldsymbol{W_b}$, the form of the decision function  \(\Phi(\boldsymbol{x})\) can be formulated as follows:
\begin{equation}
\label{eq:B_x}
\begin{aligned}
\Phi(\boldsymbol{x})={\boldsymbol{\tilde{x}}}^\top\boldsymbol{W}_b,
\end{aligned}
\end{equation}
where the predicted label of a samples \(\boldsymbol{\tilde{x}}\) is \(\tilde{y}=\arg \max\limits_i\Phi(\boldsymbol{x})\).

\subsection{Incremental Update of Lite-RVFL in the Online Stage}
Assume that after the offline training phase with \( N \) samples, the first sample in the online phase corresponds to \( t = N+1 \). Let \( \boldsymbol{A}_n \) and \( \boldsymbol{S}_n \) represent the data matrix and the feature matrix at time \( t = n \), respectively. The sample weight matrix is denoted as \( \boldsymbol{T}_n \), and the classifier output weights are represented by \( \boldsymbol{W}_b^n \). At time \( t = n+1 \), the corresponding matrices are \( \boldsymbol{A}_{n+1} \), \( \boldsymbol{S}_{n+1} \), \( \boldsymbol{T}_{n+1} \), and the updated classifier output weights are \( \boldsymbol{W}_b^{n+1} \). \( \boldsymbol{A}_{n+1} \) and \( \boldsymbol{A}_n \), as well as \( \boldsymbol{S}_{n+1} \) and \( \boldsymbol{S}_n \), satisfy the following relationships:
\begin{equation}
\label{eq:DeltaA_DeltaS}
\boldsymbol{A}_{n+1}=\begin{bmatrix}
    \boldsymbol{A}_n \\
    \Delta \boldsymbol{A}
\end{bmatrix}, \boldsymbol{S}_{n+1}=\begin{bmatrix}
    \boldsymbol{S}_n \\
    \Delta \boldsymbol{S}
\end{bmatrix}.
\end{equation}

We update \(T_{n+1}\) as follows to assign the highest weight to the most recent sample:
\begin{equation}    \boldsymbol{T}_{n+1}= \begin{bmatrix}
\boldsymbol{T}_n & 0 \\
0 & \theta^n
\end{bmatrix}.
\end{equation}

\begin{theorem}
\label{thm:update}
 \( \boldsymbol{W}_b^{n+1} \) has the following update rule relative to \( \boldsymbol{W}_b^n \):
\begin{equation}
\begin{aligned}
\boldsymbol{W}_{b}^{n+1}=\boldsymbol{W}_{b}^{n}+\boldsymbol{P}_n\Delta\boldsymbol{Q}-\Delta\boldsymbol{P}\boldsymbol{Q}_n-\Delta\boldsymbol{P}\Delta\boldsymbol{Q},\\
\ 
\end{aligned}
\end{equation}
where \[
\boldsymbol{P}_n = (\lambda\boldsymbol{I} + \boldsymbol{A}_n^\top\boldsymbol{T}_n^\top\boldsymbol{T}_n\boldsymbol{A}_n)^{-1},\]
\[\boldsymbol{Q}_n = \boldsymbol{A}_n^\top\boldsymbol{T}_n^\top\boldsymbol{T}_n\boldsymbol{S}_n ,
\]
\[
\Delta\boldsymbol{P} = \theta^{2 n} \boldsymbol{P}_n \boldsymbol{\Delta A}^{\top}\left(1+\theta^{2 n} \boldsymbol{\Delta A} \boldsymbol{P}_n \boldsymbol{\Delta A}^{\top}\right)^{-1} \boldsymbol{\Delta A} \boldsymbol{P}_n, \]
\[\Delta\boldsymbol{Q} = \theta^{2n}\boldsymbol{\Delta A}^\top \Delta \boldsymbol{S}.
\]
\end{theorem}
\begin{pf}
According to Eqs.~(\ref{eq:W_b})(\ref{eq:DeltaA_DeltaS}), the weight matrix \(\boldsymbol{W}_b^{n+1}\) at time \(t=n+1\) can be expressed as:
\begin{equation}
\label{eq:W_b_2}
\begin{aligned}
\boldsymbol{W}_{b}^{n+1}= &\underbrace{\left(\lambda \boldsymbol{I}+\boldsymbol{{A}}_{n}^\top\boldsymbol{T}_{n}^\top\boldsymbol{T}_{n} \boldsymbol{{A}}_{n}+\theta^{2n}\boldsymbol{\Delta A}^\top\boldsymbol{\Delta A}\right)^{-1}}_{\boldsymbol{P}_{n+1}} \cdot \\
&\underbrace{\left(\boldsymbol{{A}}^\top\boldsymbol{T}_n^\top\boldsymbol{T}_n \boldsymbol{{S}_n}+\theta^{2n}\boldsymbol{\Delta A}^\top \Delta \boldsymbol{S}\right)}_{\boldsymbol{Q}_{n+1}}.
\end{aligned}
\end{equation}

Denote \(\Delta \boldsymbol{M}=\left(1+\theta^{2 n} \boldsymbol{\Delta A} \boldsymbol{P}_n \boldsymbol{\Delta A}^{\top}\right)^{-1}\). Based on Woodbury matrix identity~\cite{hager1989updating}, \(\boldsymbol{P}_{n+1}\) and \(\boldsymbol{Q}_{n+1}\) in Eq.~(\ref{eq:W_b_2}) can be transformed into:
\begin{equation}
\label{eq:K_t+1}
\boldsymbol{P}_{n+1}=\boldsymbol{P}_n-\underbrace{\theta^{2 n} \boldsymbol{P}_n \boldsymbol{\Delta A}^{\top}\Delta \boldsymbol{M} \boldsymbol{\Delta A} \boldsymbol{P}_n}_{\Delta\boldsymbol{P}},
\end{equation}
and
\begin{equation}
\label{eq:Q_t+1}
\boldsymbol{Q}_{n+1}=\boldsymbol{Q}_n+\underbrace{\theta^{2n}\Delta\boldsymbol{A}^\top\Delta\boldsymbol{S}}_{\Delta\boldsymbol{Q}}.
\end{equation}
Since \( \boldsymbol{W}_b^n = \boldsymbol{P}_n \boldsymbol{Q}_n \), the proof is complete.
\end{pf}
\begin{theorem}
\label{thm:constant}
   When \( t = n \) becomes sufficiently large, the proportion of the nearest \( L \) samples approaches a constant value, given by \( 1 - \theta^{-L} \).
\end{theorem}
\begin{pf}
The total weight of all samples, denoted by \( W_{\text{all}} \), is given by the sum of the geometric series:
\begin{equation}
W_{\text{all}} = \sum_{i=1}^n \theta^{i-1} = \frac{\theta^n - 1}{\theta - 1}.
\end{equation}
Similarly, the weight of the nearest \( L \) samples, \( W_L \), is expressed as:
\begin{equation}
W_L = \sum_{i=n-L+1}^n \theta^{i-1} = \frac{\theta^{n-L}(\theta^L - 1)}{\theta - 1}.
\end{equation}

The proportion \( p \) of the nearest \( L \) samples relative to the total weight is then obtained:
\begin{equation}
p = \frac{W_L}{W_{\text{all}}} = \frac{1 - \theta^{-L}}{1 - \theta^{-n}}.
\end{equation}
Taking the limit as \( n \to \infty \), we obtain:
\begin{equation}
\lim_{n \to \infty} \frac{1 - \theta^{-L}}{1 - \theta^{-n}} = 1 - \theta^{-L}.
\end{equation}

Thus, as \( N \) becomes large, the proportion \( p \) converges to the constant value \( 1 - \theta^{-L} \), completing the proof.
\end{pf}

The parameter \(\theta\) determines the relative emphasis the classifier places on new versus old concepts. A larger \(\theta\) increases the sensitivity of the classifier to newly observed concepts, thereby accelerating adaptation. The adapting speed for new concepts can thus be controlled  by adjusting \(\theta\). To achieve a contribution ratio of \(\alpha\) from the latest \(L\) samples, \(\theta\) should be set to 
\begin{equation}
\theta=(1 - \alpha)^{-1/L}. 
\end{equation}

For instance, to ensure that the most recent 500 samples contribute 80\% of the influence on the classifier, the condition \(1 - \theta^{-500} = 80\%\) can be used, yielding \(\theta = 1.003\). Notably, this same value of $\theta$ results in the most recent 200 samples contributing approximately 50\% of the influence. This implies that setting \(\theta = 1.003\) enables the classifier to almost fully adapt to concept drift within 200 to 500 samples.

\subsection{Comparison with an Alternative Method}
To highlight the effectiveness of Lite-RVFL, an alternative method named Alt-RVFL is introduced in this subsection, which also assigns higher weights to more recent samples but fails to adapt effectively to drift. Specifically, it employs weights based on powers of natural numbers. In this case, \( \boldsymbol{T}_N \) and the updates for \( \boldsymbol{T}_n \) are as follows:

\begin{equation}
\label{eq:TN_2}
\boldsymbol{T}_N = 
\begin{bmatrix}
1 & 0 & \cdots & 0 \\
0 & 2^k & \cdots & 0 \\
\vdots & \vdots & \ddots & \vdots \\
0 & 0 & \cdots & N^k \\
\end{bmatrix},
\boldsymbol{T}_{n+1}=
\begin{bmatrix}
\boldsymbol{T}_n & 0 \\
0 & (n+1)^k
\end{bmatrix},
\end{equation}
where \( k \) is a positive integer.

Alt-RVFL can still perform incremental updates, as derived similarly to Theorem~\ref{thm:update}. However, this classifier no longer satisfies the conclusion presented in Theorem~\ref{thm:constant}, where the proportion of the newest \( L \) samples remains constant when \(t\) is large enough, as demonstrated in Corollary~\ref{coro:compared}.
\begin{coro}
\label{coro:compared}
When the classifier uses the weight matrix and the update rule in Eq.~(\ref{eq:TN_2}), as \( t = n \) becomes sufficiently large, the proportion of the nearest \( L \) samples tends to 0.
\end{coro}
\begin{pf}
According to the Sum of Powers Formula \cite{bounebirat2017several, si2019powers}, the total weight of all samples, denoted by \( W_{\text{all}} \), is given by the following expression:
\begin{equation}
W_{\text{all}} = \sum_{i=1}^n i^{k} =\frac{n^{k+1}}{k+1}+\frac{B_k}{2} n^k+O\left(n^{k-1}\right) .
\end{equation}
Similarly, the weight of the nearest \( L \) samples, \( W_L \), is expressed as:
\begin{equation}
W_L = W_{\text{all}}-\frac{(n-L)^{k+1}}{k+1}-\frac{B_k}{2} (n-L)^k.
\end{equation}
It is evident that as \( n \) tends to infinity, \( {W_L}/{W_{\text{all}}} = 0 \).
\end{pf}

Therefore, under this alternative approach, when \( n \) becomes sufficiently large, the contribution of the nearest \( L \) samples to the classifier becomes zero. If concept drift occurs at this point, \( L \) must be on the same order of magnitude as \( n \) to adapt to the drift, which would require a considerable amount of time.

\section{Experiments}
\label{sec:experiments}
\begin{figure}[!h]
    \centering

    \includegraphics[width=\linewidth]{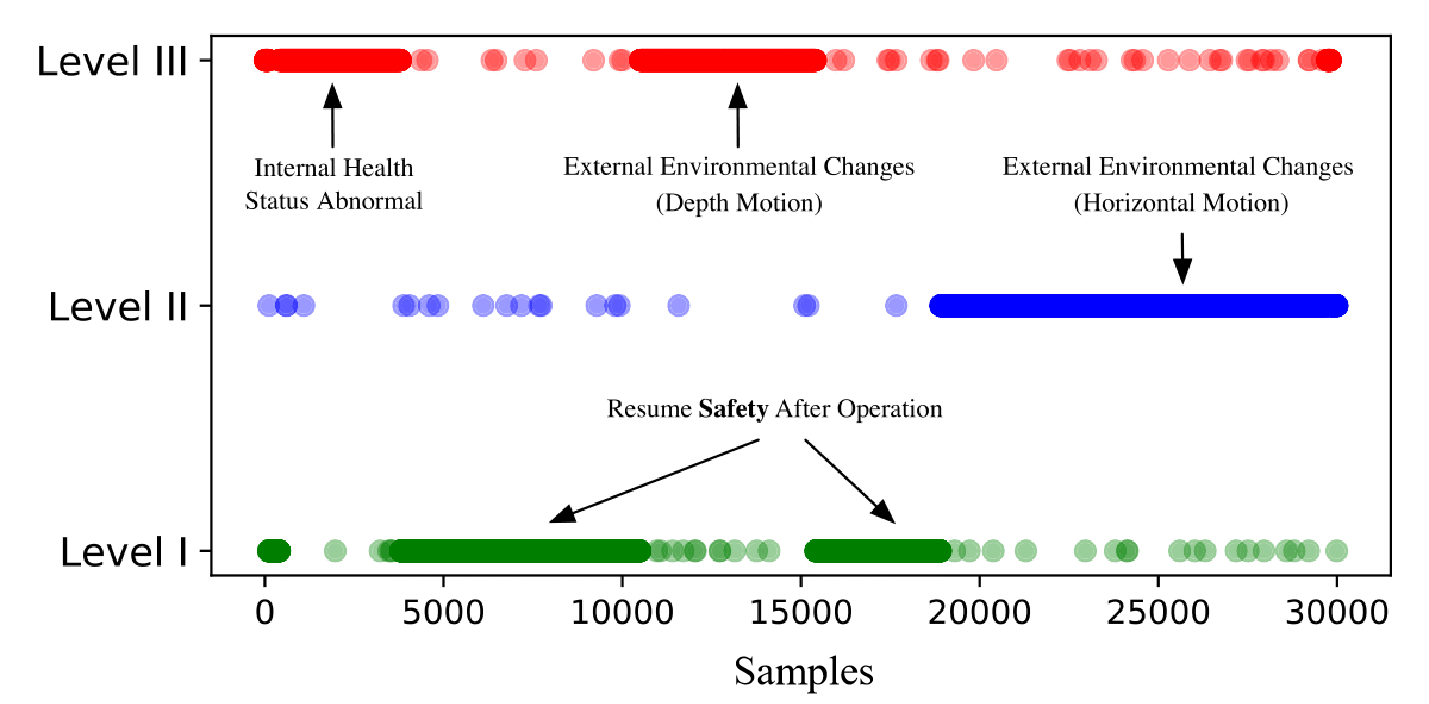}

        \caption{Illustration of the DSMS dataset.}
        \label{fig:dataset}
\end{figure}

In this section, the effectiveness of the proposed method is validated through a practical safety assessment task. The dataset used is derived from the Deep-sea Manned Submersible (DSMS), specifically the exploration task data from March 19, 2017 \cite{hu2024cadm}. This dataset, sourced from the life support system, comprises 24 features from various sensors, including carbon dioxide concentration, oxygen concentration, posture angles, thrust, and moment. As illustrated in Fig.~\ref{fig:dataset}, the data is categorized into three safety levels: safe, generally safe, and unsafe. The dataset contains a total of 30,000 samples, with 8,514 samples classified as safe, 10,974 as generally safe, and 10,512 as unsafe\footnote{For more details, please refer to the website: \url{https://github.com/THUFDD/JiaolongDSMS\_datasets}.}. Both real and virtual concept drifts are included in this dataset, as the criteria for evaluating safety vary depending on the depth. All experiments are implemented in Python on a platform equipped with an Intel i5-13600KF CPU, boasting 14 cores, a 3.50-GHz clock speed, and 20 processors, complemented by 32 GB of RAM.

\subsection{Settings}
\subsubsection{Task Description} The first 200 samples of the dataset are used as an offline training set, while the remaining 29,800 samples arrive sequentially in the form of a data stream. The classifier is required to make predictions upon the arrival of each sample. After making a prediction, the classifier receives the true label of the sample and updates itself. 
\subsubsection{Evaluation Metrics}  
Accuracy is the primary consideration, while runtime is also taken into account.
\subsubsection{Method Configuration} The number of node groups in the RVFL is set to 10, with 10 nodes in each group. The activation function used is the sigmoid function. Additionally, \(\lambda = 0.1\) and \(\theta = 1.003\) are employed.
\subsubsection{Comparative Methods} The proposed method is compared with a set of hybrid models combining RVFL with benchmark concept drift detectors, including RVFL-ADWIN \cite{bifet2007learning}, RVFL-HDDMw \cite{frias2014online}, RVFL-HDDMa \cite{frias2014online}, and RVFL-PageHinkley \cite{page1954continuous}. These drift detection algorithms detect concept drift by monitoring changes in data statistics over time using adaptive windowing or statistical analysis of cumulative deviations. Once drift is detected, the classifiers of these methods are retrained using the most recent 200 samples. The implementations are based on the default configurations provided by the scikit-multiflow library \footnote{\url{https://scikit-multiflow.github.io}.}. Furthermore, to verify the unique role of the theoretical property proposed in Theorem~\ref{thm:constant} in adapting to concept drift, we also compare against the proposed Alt-RVFL where \(k\) is set to 2. All comparative methods adopt the same RVFL settings as used in our approach.

\subsection{Results}
\begin{figure}[!h]
    \centering

    \includegraphics[width=0.89\linewidth]{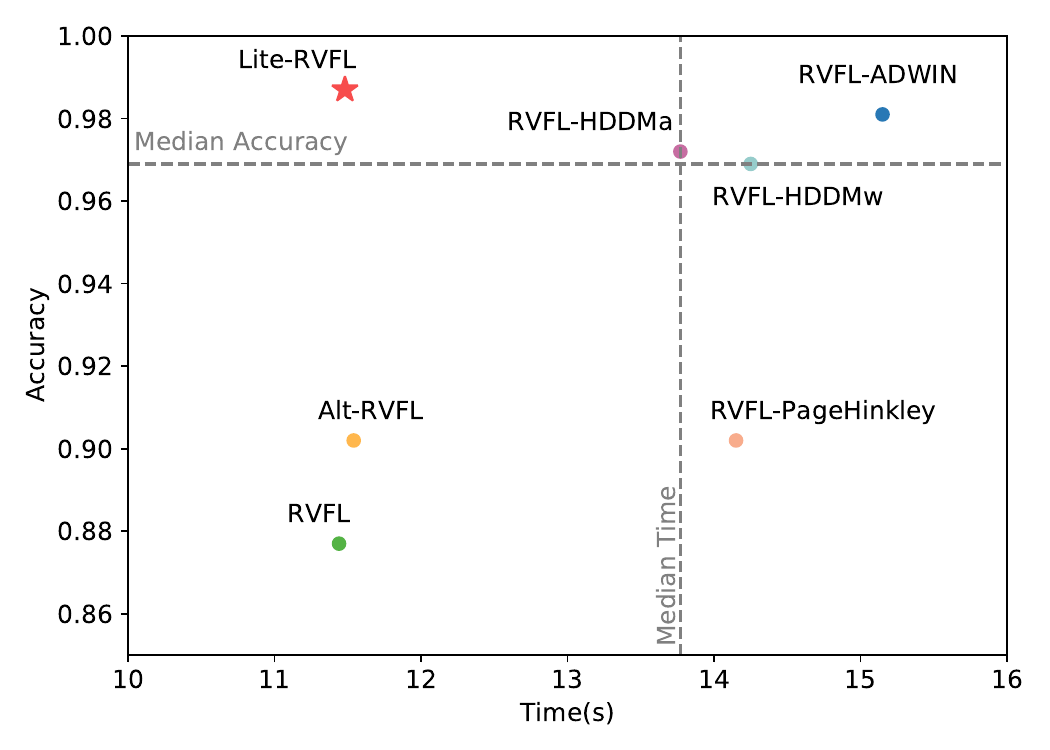}

        \caption{Average runtime and accuracy of different methods.}
        \label{fig:time_acc}
\end{figure}

\begin{figure}[h]
    \centering

    \includegraphics[width=0.89\linewidth]{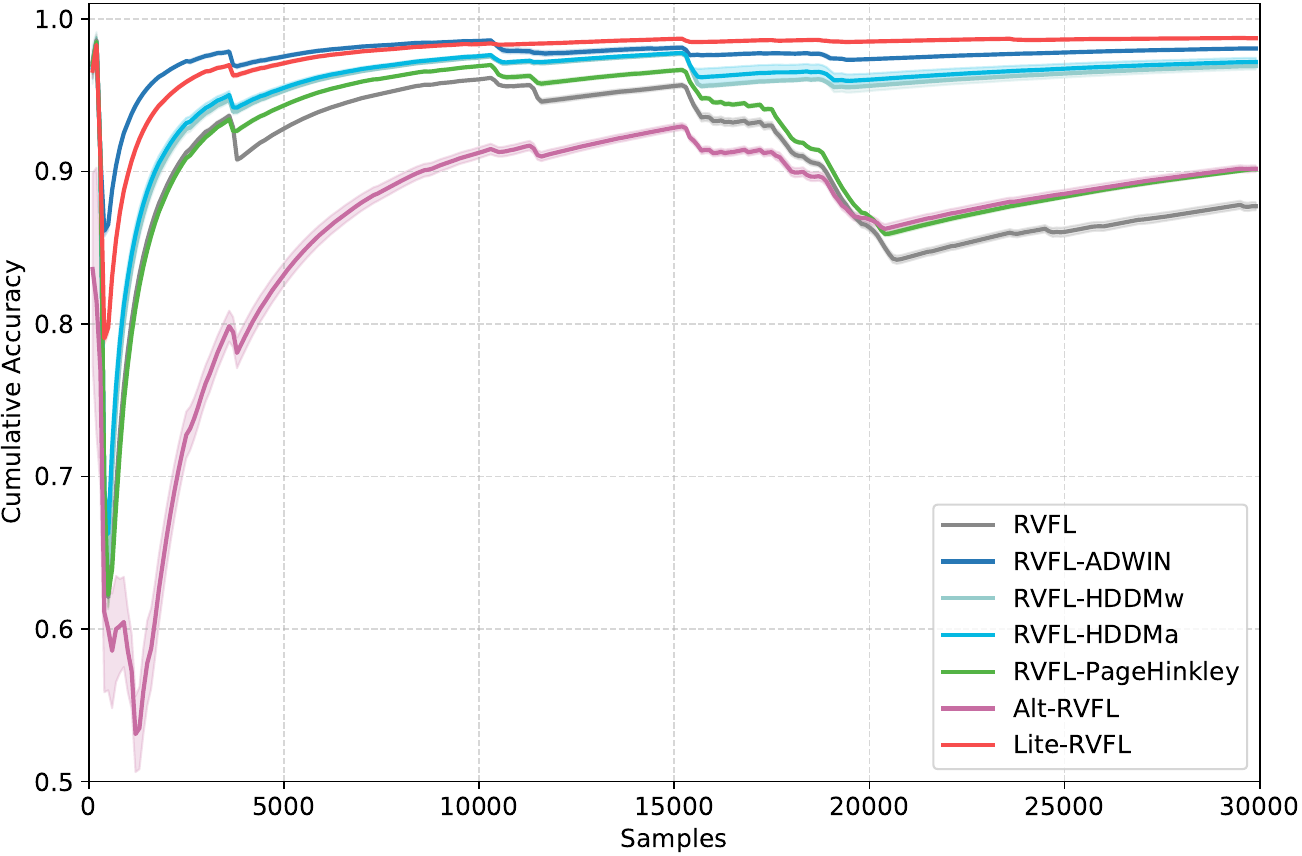}

        \caption{The learning curves of different methods on DSMS dataset with concept drift. {The shaded areas represent the standard deviation results using the corresponding method under multiple random tests.}}
        \label{fig:acc_comparison}
\end{figure}

\begin{figure*}[htbp]
    \centering

    \begin{subfigure}[c]{0.33\linewidth}
        \centering
        \includegraphics[width=\linewidth]{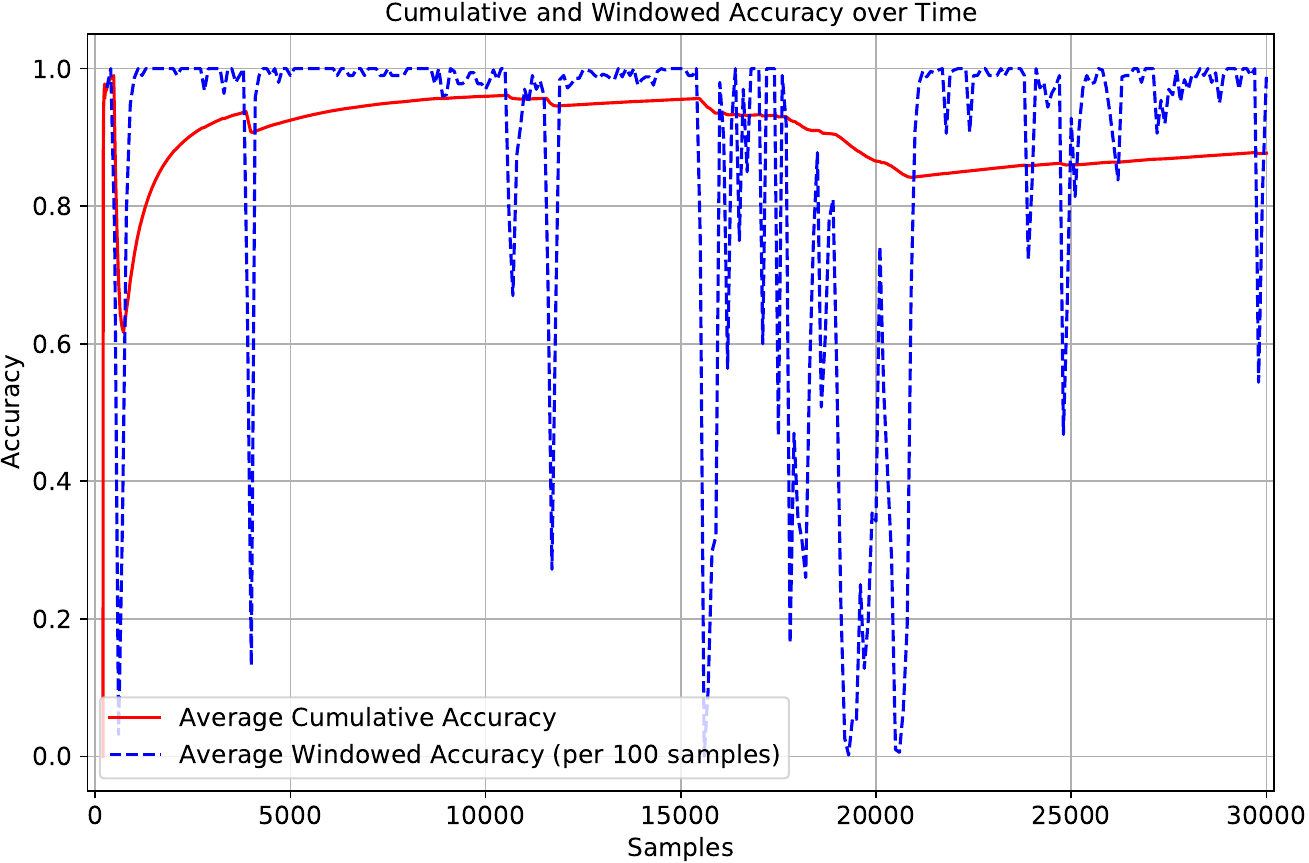}
        \caption{RVFL}
    \end{subfigure}%
    \hfill
    \begin{subfigure}[c]{0.33\linewidth}
        \centering
        \includegraphics[width=\linewidth]{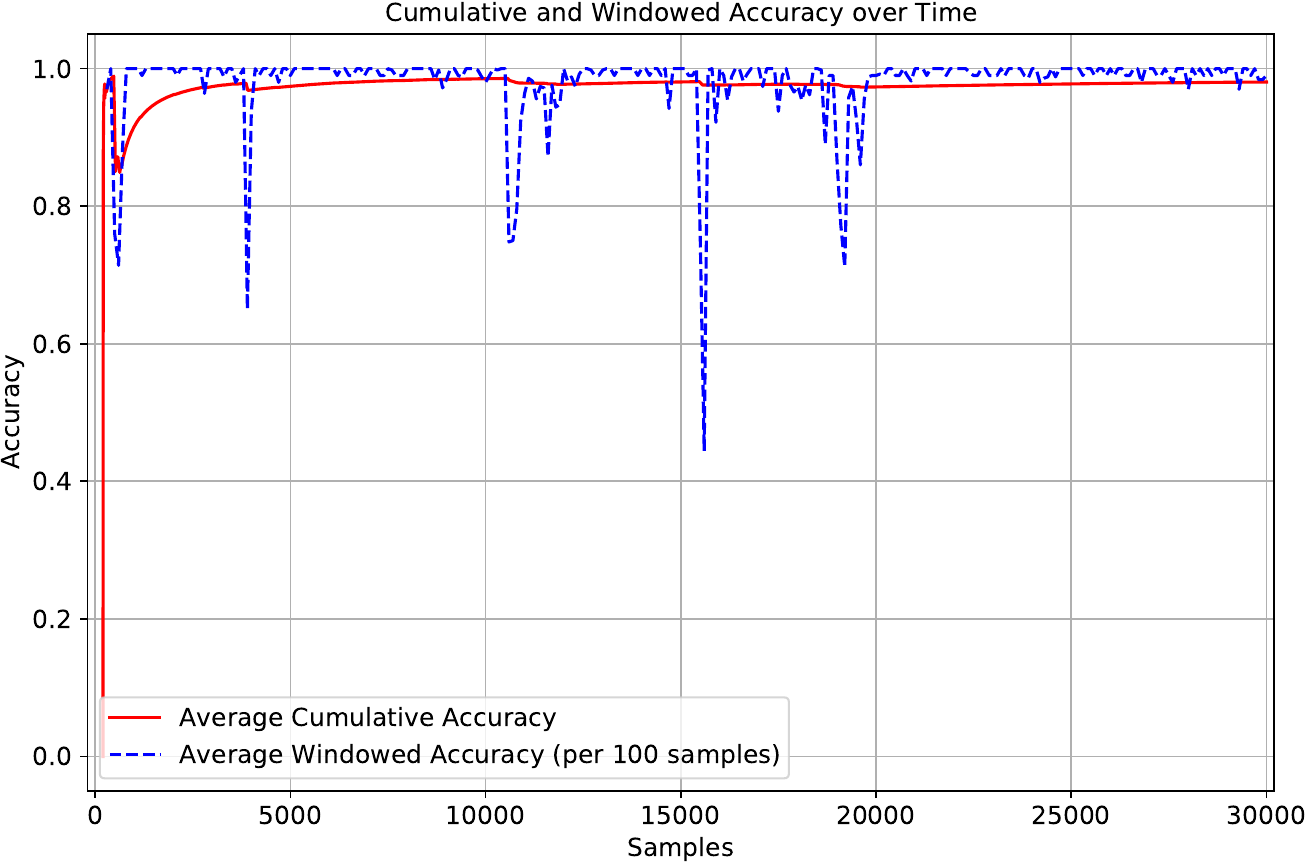}
        \caption{RVFL-ADWIN}
    \end{subfigure}%
    \hfill
    \begin{subfigure}[c]{0.33\linewidth}
        \centering
        \includegraphics[width=\linewidth]{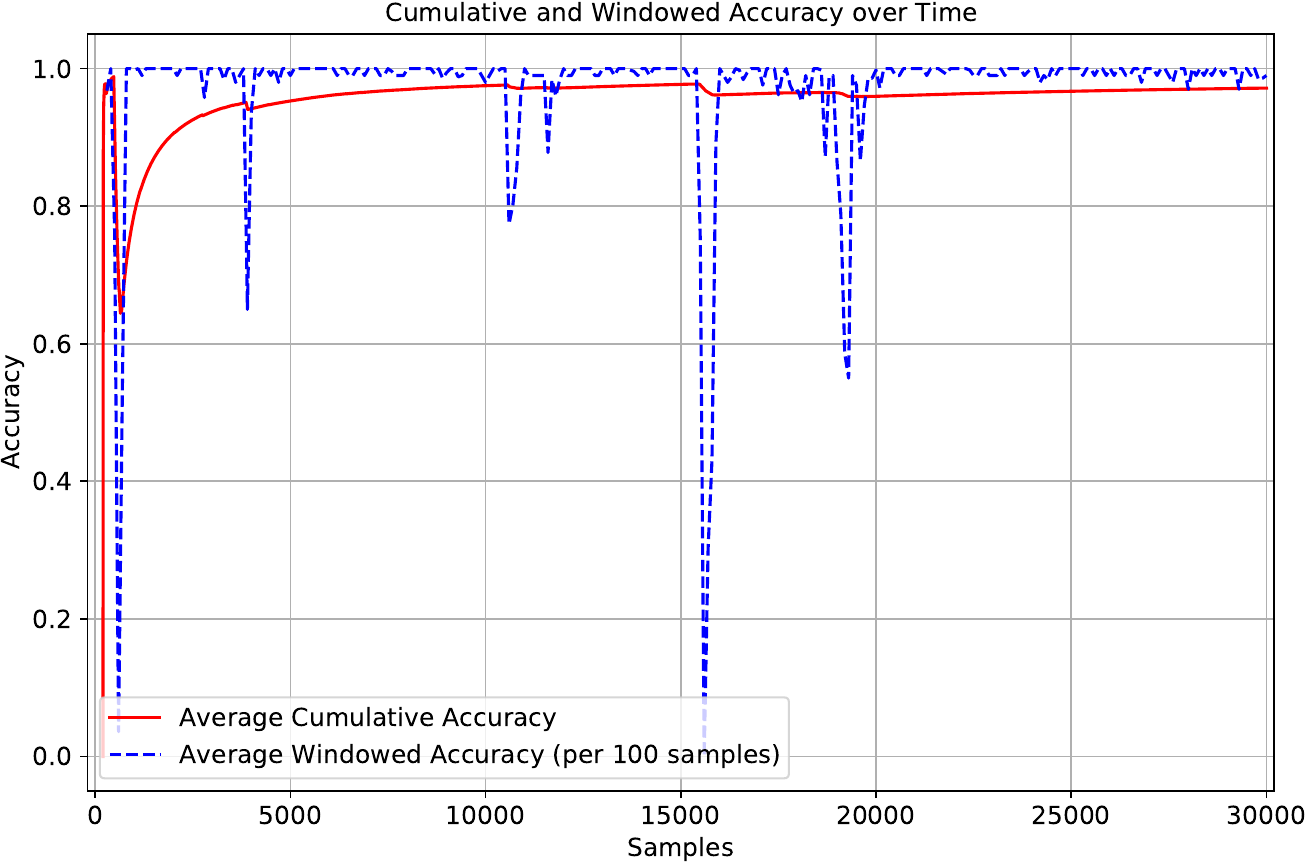}
        \caption{RVFL-HDDM\textsubscript{a}}
    \end{subfigure}

    \vskip\baselineskip

    \begin{subfigure}[c]{0.33\linewidth}
        \centering
        \includegraphics[width=\linewidth]{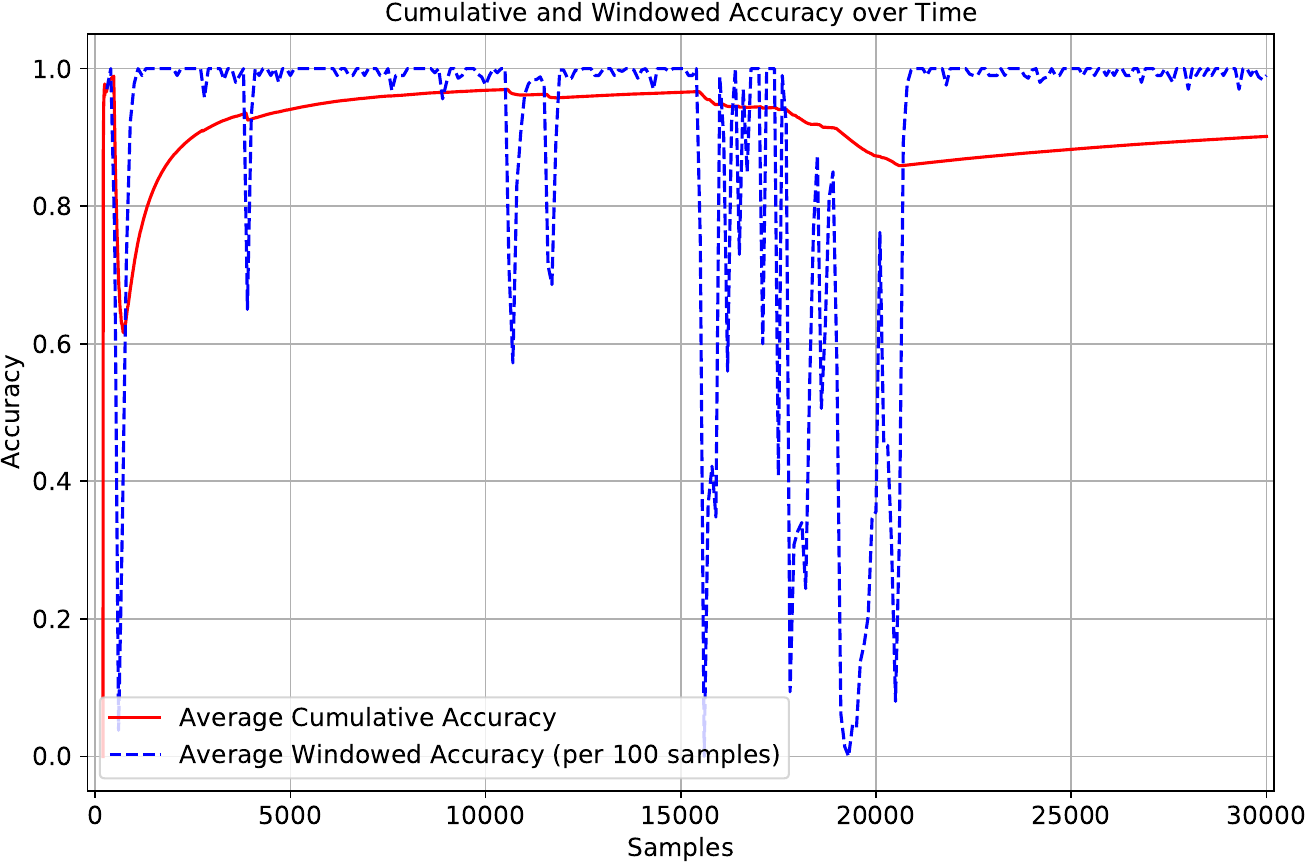}
        \caption{RVFL-PageHinkley}
    \end{subfigure}%
    \hfill
    \begin{subfigure}[c]{0.33\linewidth}
        \centering
        \includegraphics[width=\linewidth]{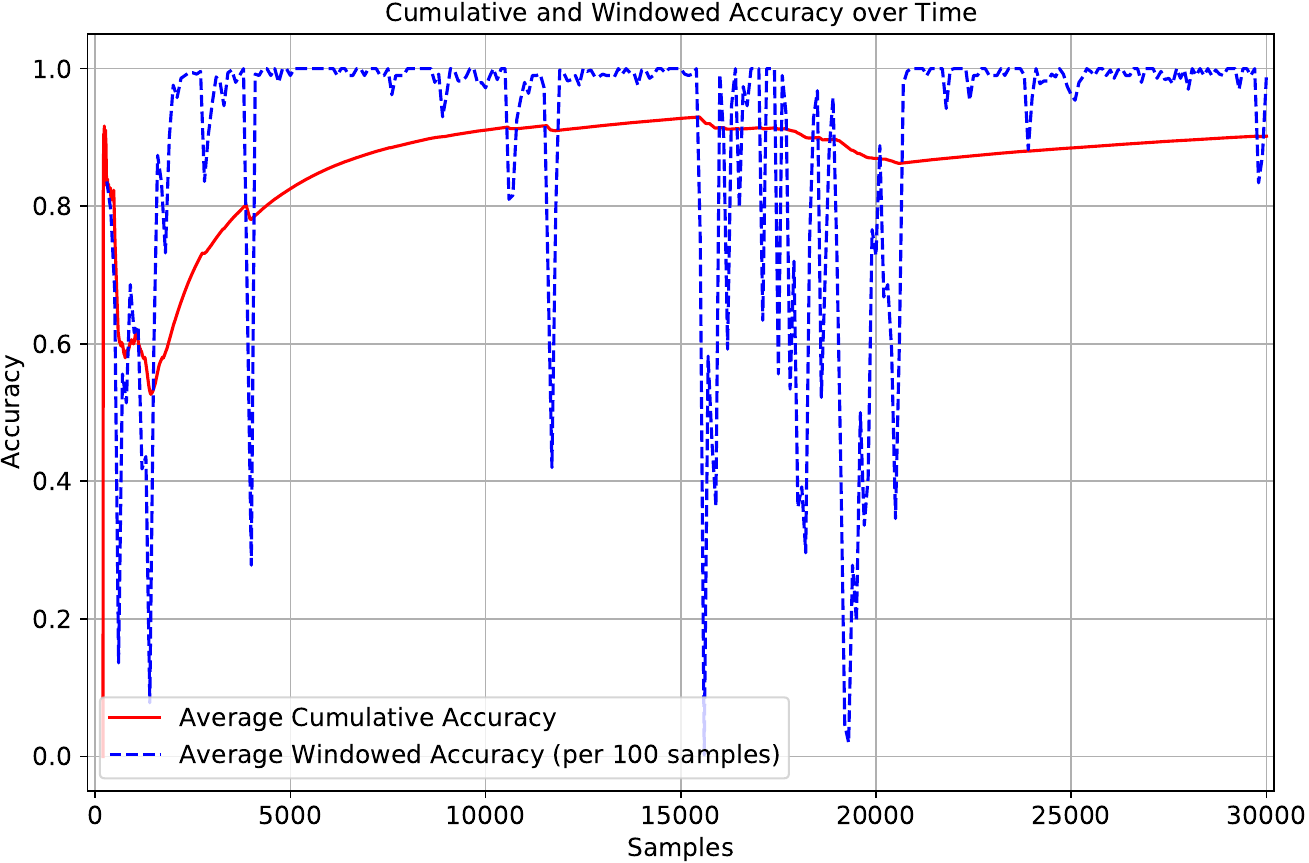}
        \caption{Alt-RVFL}
    \end{subfigure}%
    \hfill
    \begin{subfigure}[c]{0.33\linewidth}
        \centering
        \includegraphics[width=\linewidth]{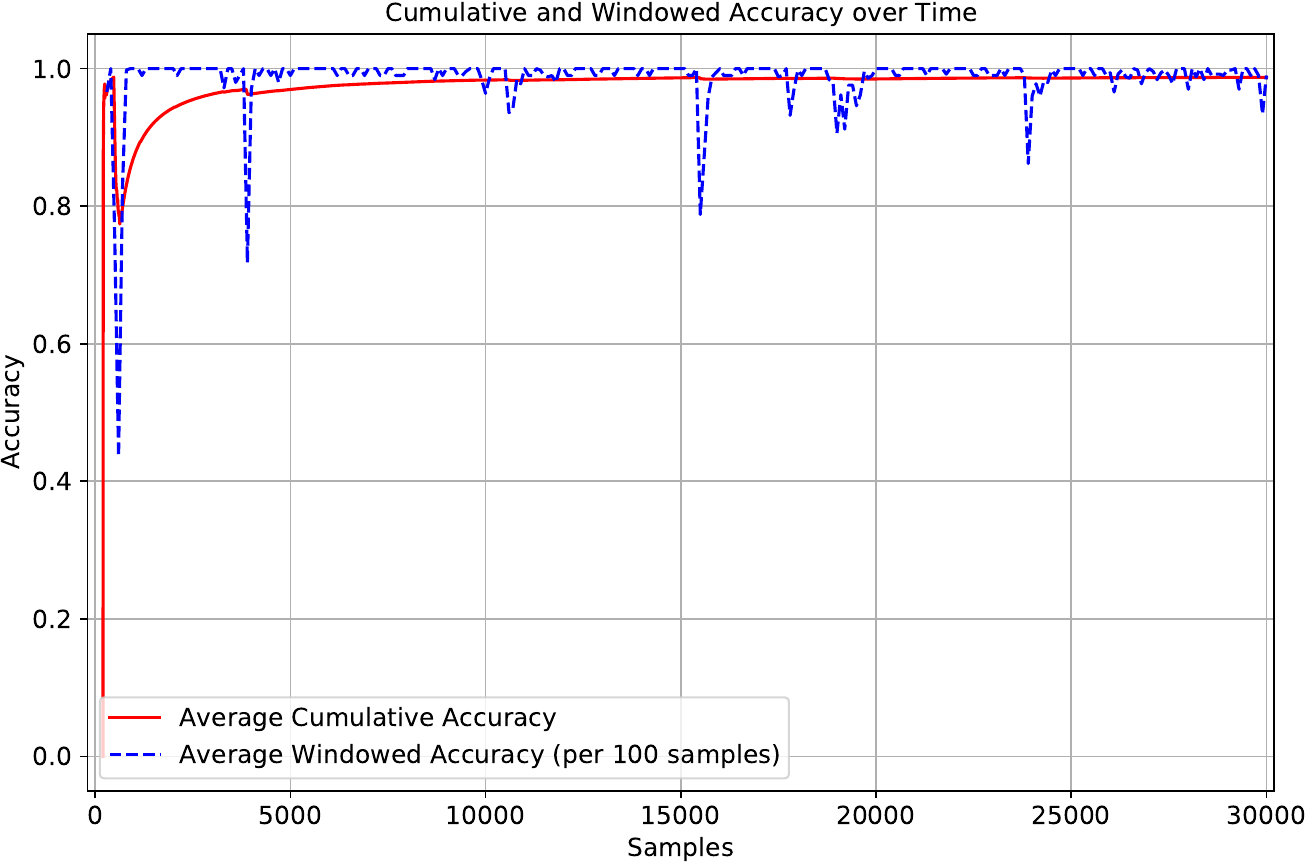}
        \caption{Lite-RVFL (Proposed)}
    \end{subfigure}

    \caption{Cumulative and windowed accuracy for different methods.}
    \label{fig:cul_win}
\end{figure*}

In this subsection, we present a comparison with several benchmark methods. The results are summarized in Table~\ref{tab:results} and visualized in Figs.~\ref{fig:time_acc}-\ref{fig:cul_win}.


\begin{table*}[h]
\centering
\caption{Accuracy and time (mean $\pm$ std) of different methods over 5 runs.}
\label{tab:algorithm_comparison}
\resizebox{\linewidth}{!}{
\begin{tabular}{c c c c c c c c} 
\specialrule{0.15em}{1pt}{1pt}
\specialrule{0.15em}{1pt}{3pt}

{\bf Methods} 
 & \bf RVFL & \bf RVFL-ADWIN 
 & \bf RVFL-HDDMw 
 & \bf RVFL-HDDMa
 & \bf RVFL-PageHinkley 
 & \bf Alt-RVFL 
 & \bf Lite-RVFL* \\
\specialrule{0.15em}{1pt}{3pt}

{\bf Accuracy}  
 & 87.71\% \(\pm\) 0.27\%
 &\textbf{98.06\% \(\pm\) 0.04\%} 
 & 96.90\% \(\pm\) 0.22\% 
 & 97.17\% \(\pm\) 0.27\% 
 & 90.15\% \(\pm\) 0.11\%  
 & 90.18\% \(\pm\) 0.22\% 
 & \textbf{98.73\% \(\pm\) 0.03\%} \\ 
\specialrule{0.1em}{3pt}{3pt}
 
{\bf Rank} 
 & 7
 & \textbf{2}  
 & 4 
 & 3 
 & 6  
 & 5 
 & \textbf{1} \\ 
\specialrule{0.1em}{3pt}{3pt}

{\bf Time (s)}   
 &  \textbf{11.44 \(\pm\) 0.15}
 & {15.15 \(\pm\) 0.08}
 &14.25 \(\pm\) 0.13
 & 13.77 \(\pm\) 0.36 
 & 14.15 \(\pm\) 0.17 
 & 11.54 \(\pm\) 0.18 
 & \textbf{11.48 \(\pm\) 0.07} \\ 

\specialrule{0.10em}{3pt}{3pt}

{\bf Rank} 
 & \textbf{1}
 & 7  
 & 6 
 & 4 
 & 5  
 & 3 
 & \textbf{2} \\ 

\specialrule{0.10em}{3pt}{3pt}

 {\bf Drifts Detected} 
 & 0
 & 22  
 & 5 
 & 5 
 & 2  
 & ------ 
 & ------ \\ 

\specialrule{0.15em}{3pt}{1pt}
\specialrule{0.15em}{1pt}{1pt}

\end{tabular}}
\begin{tablenotes}\footnotesize
    \item[1] *Note 1: The notation `*' represents the proposed approach. The notation ‘------’ indicates that the corresponding method does not require drift detection.
    \item[2] *Note 2: The Top-1 and Top-2 performances for each dataset are \textbf{bolded} in the table.     
\end{tablenotes}
\label{tab:results}
\end{table*}

Due to the presence of concept drift, the RVFL, lacking an effective drift handling mechanism, experiences a sharp decline in performance after a drift occurs, with a prolonged recovery period. As shown in Fig.~\ref{fig:cul_win}(a), two minor drifts occur around the 4,000th and 10,000th samples, while a more significant drift occurs at the 15,000th sample. Each time a drift occurs, the windowed accuracy of the RVFL suddenly drops and takes a considerable amount of time to recover through incremental updates. As a result, it achieves an accuracy of only 87.71\%. However, due to its simplicity, the RVFL operates with the fastest runtime, completing the process in just 11.44 seconds. For RVFL-ADWIN, it quickly detects the three drifts and performs retraining using the most recent samples to rapidly adapt to the changes. Additionally, it detects and handles drift during periods of minor accuracy drops. As a result, it achieves a high overall accuracy of 98.06\%, as shown in Fig.~\ref{fig:cul_win}(b). However, its sensitive drift detection mechanism identifies a total of 22 drifts, meaning the classifier undergoes retraining 22 times. Consequently, it has the slowest runtime, taking 15.15 seconds. For RVFL-HDDMw and RVFL-HDDMa, both detect five drifts, including the three major drifts, resulting in relatively high accuracy. Furthermore, their runtimes are shorter than that of RVFL-ADWIN, as shown in Fig.~\ref{fig:cul_win}(c). In contrast, RVFL-PageHinkley detects only two drifts, excluding the one around the 15,000th sample. As a result, the accuracy during this period is significantly lower, leading to an overall accuracy of only 90.18\%, as shown in Fig.~\ref{fig:cul_win}(d).

It is evident that, in the presence of concept drift, the performance of the classifier heavily depends on the drift detection. Excessively sensitive drift detection can lead to high time consumption due to frequent retraining. On the other hand, overly conservative drift detection may fail to detect critical drifts, causing a significant decline in classifier performance. The classifier we propose, Lite-RVFL, assigns higher weights to each new sample and ensures that the most recent \(L\) samples make the primary contribution to the decision of the classifier. This mechanism enables automatic adaptation to new concepts during incremental updates, eliminating the need for additional drift detection and retraining. Moreover, this approach better captures temporal relationships, leading to a higher accuracy of 98.73\%, surpassing that of RVFL-ADWIN. Since this mechanism is integrated into the classifier structure, it introduces no additional computational overhead, and its runtime is nearly identical to that of RVFL, taking 11.48 seconds, as shown in Fig.~\ref{fig:cul_win}(f). However, not all such weight designs achieve the desired effect. For example, the alternative method we proposed fails to ensure that the most recent \(L\) samples have the main influence on the classifier. As a result, it cannot quickly adapt after a drift occurs, and its performance is only slightly better than that of RVFL, as shown in Fig.~\ref{fig:cul_win}(e). 

\subsection{Discussion}
Lite-RVFL achieves concept drift adaptation without the need for explicit drift detection or retraining, relying solely on its internal structure. However, since the weights assigned to new samples increase exponentially over time, the  influence of the regularization term \(\lambda \left\|\boldsymbol{W}_b\right\|_2^2\) gradually diminishes as the number of samples grows. This may lead to overfitting, particularly on datasets with low dimensionality and a small number of classes, where the classifier focuses excessively on the most recent 
\(L\) samples, thereby reducing its generalization ability to new data and ultimately degrading accuracy.

To mitigate this phenomenon, several solutions can be considered: 1) Periodically retraining the classifier after a large number of samples (e.g., every 5000 samples) to prevent the accumulation of weights of samples relative to \(\lambda \left\|\boldsymbol{W}_b\right\|_2^2\); 2) Increasing the regularization coefficient \(\lambda\) or decreasing the forgetting factor \(\theta\), although such adjustments may degrade generalization performance or hinder the ability to adapt to concept drift; 3) Introducing a smaller forgetting factor \(\theta' < 1\) instead of \(\theta>1\), and modifying Eq.~\eqref{eq:TN_2} as follows:

\begin{equation}
\label{eq:TN_3}
\boldsymbol{T}_N = 
\begin{bmatrix}
\theta'^{N-1} & 0 & \cdots & 0 \\
0 & \theta'^{N-2} & \cdots & 0 \\
\vdots & \vdots & \ddots & \vdots \\
0 & 0 & \cdots & 1 \\
\end{bmatrix},
\boldsymbol{T}_{n+1}=
\begin{bmatrix}
\theta'\boldsymbol{T}_n & 0 \\
0 & 1
\end{bmatrix},
\end{equation}
which can explicitly control the overall weights of samples, but makes it difficult to derive a closed-form solution for incremental updates; and 4) Reformulating the current objective function in Eq.~\eqref{eq:optimization}, which is essentially designed for regression, into a form that directly targets classification tasks. 

While these solutions demonstrate  promise in addressing the overfitting problem, their effectiveness require more comprehensive investigation and validation.

\section{Conclusion}
\label{sec:conclusion}
This paper has presented {Lite-RVFL}, a lightweight RVFL for learning under concept drift. Lite-RVFL introduces exponentially inceasing weightes to new samples in the  objective function that enables the model to focus on recent samples. Theoretical analysis has demonstrated that the Lite-RVFL can maintain stable attention to recent samples, thus supporting its effectiveness in handling drifts. Furthermore, an efficient incremental update formulation has been derived. Experiments have been conducted on a real-world RTSA task of the Jiaolong deep-sea manned submersible, which have confirmed that Lite-RVFL outperforms RVFL models combined with drift detectors in terms of accuracy, while maintaining computational efficiency comparable to that of the standard RVFL. These results highlight the potential of Lite-RVFL as a fast and adaptive solution for online learning tasks in non-stationary environments. In the future, we will investigate solutions to the overfitting problem that may arise in low-dimensional data. Additionally, we aim to extend Lite-RVFL to an ensemble version with theoretical performance bounds.

\bibliographystyle{IEEEtran}
\bibliography{References}

\begin{thebibliography}{10}
\providecommand{\url}[1]{#1}
\csname url@samestyle\endcsname
\providecommand{\newblock}{\relax}
\providecommand{\bibinfo}[2]{#2}
\providecommand{\BIBentrySTDinterwordspacing}{\spaceskip=0pt\relax}
\providecommand{\BIBentryALTinterwordstretchfactor}{4}
\providecommand{\BIBentryALTinterwordspacing}{\spaceskip=\fontdimen2\font plus
\BIBentryALTinterwordstretchfactor\fontdimen3\font minus
  \fontdimen4\font\relax}
\providecommand{\BIBforeignlanguage}[2]{{%
\expandafter\ifx\csname l@#1\endcsname\relax
\typeout{** WARNING: IEEEtran.bst: No hyphenation pattern has been}%
\typeout{** loaded for the language `#1'. Using the pattern for}%
\typeout{** the default language instead.}%
\else
\language=\csname l@#1\endcsname
\fi
#2}}
\providecommand{\BIBdecl}{\relax}
\BIBdecl

\bibitem{hoi2021online}
S.~C. Hoi, D.~Sahoo, J.~Lu, and P.~Zhao, ``Online learning: A comprehensive
  survey,'' \emph{Neurocomputing}, vol. 459, pp. 249--289, 2021.

\bibitem{lu2018learning}
J.~Lu, A.~Liu, F.~Dong, F.~Gu, J.~Gama, and G.~Zhang, ``Learning under concept
  drift: A review,'' \emph{IEEE transactions on knowledge and data
  engineering}, vol.~31, no.~12, pp. 2346--2363, 2018.

\bibitem{liu2022online}
Z.~Liu, Y.~Zhang, Z.~Ding, and X.~He, ``An online active broad learning
  approach for real-time safety assessment of dynamic systems in nonstationary
  environments,'' \emph{IEEE Transactions on Neural Networks and Learning
  Systems}, vol.~34, no.~10, pp. 6714--6724, 2022.

\bibitem{liu2023real}
Z.~Liu, S.~Hu, and X.~He, ``Real-time safety assessment of dynamic systems in
  non-stationary environments: A review of methods and techniques,'' in
  \emph{2023 CAA Symposium on Fault Detection, Supervision and Safety for
  Technical Processes (SAFEPROCESS)}.\hskip 1em plus 0.5em minus 0.4em\relax
  IEEE, 2023, pp. 1--6.

\bibitem{vzliobaite2016overview}
I.~{\v{Z}}liobait{\.e}, M.~Pechenizkiy, and J.~Gama, ``An overview of concept
  drift applications,'' \emph{Big data analysis: new algorithms for a new
  society}, pp. 91--114, 2016.

\bibitem{li2025dynamic}
W.~Li, Z.~Liu, P.~Han, X.~He, L.~Wang, and T.~Zhang, ``A dynamic anchor-based
  online semi-supervised learning approach for fault diagnosis under variable
  operating conditions,'' \emph{Neurocomputing}, p. 130137, 2025.

\bibitem{han2022survey}
M.~Han, Z.~Chen, M.~Li, H.~Wu, and X.~Zhang, ``A survey of active and passive
  concept drift handling methods,'' \emph{Computational Intelligence}, vol.~38,
  no.~4, pp. 1492--1535, 2022.

\bibitem{bifet2007learning}
A.~Bifet and R.~Gavalda, ``Learning from time-changing data with adaptive
  windowing,'' in \emph{Proceedings of the 2007 SIAM international conference
  on data mining}.\hskip 1em plus 0.5em minus 0.4em\relax SIAM, 2007, pp.
  443--448.

\bibitem{frias2014online}
I.~Frias-Blanco, J.~del Campo-{\'A}vila, G.~Ramos-Jimenez, R.~Morales-Bueno,
  A.~Ortiz-Diaz, and Y.~Caballero-Mota, ``Online and non-parametric drift
  detection methods based on hoeffding’s bounds,'' \emph{IEEE Transactions on
  Knowledge and Data Engineering}, vol.~27, no.~3, pp. 810--823, 2014.

\bibitem{hu2024cadm}
S.~Hu, Z.~Liu, M.~Li, and X.~He, ``{CADM} $+ $: Confusion-based learning
  framework with drift detection and adaptation for real-time safety
  assessment,'' \emph{IEEE Transactions on Neural Networks and Learning
  Systems}, 2024.

\bibitem{zhang2020reinforcement}
H.~Zhang, W.~Liu, and Q.~Liu, ``Reinforcement online active learning ensemble
  for drifting imbalanced data streams,'' \emph{IEEE Transactions on Knowledge
  and Data Engineering}, vol.~34, no.~8, pp. 3971--3983, 2020.

\bibitem{jiao2022dynamic}
B.~Jiao, Y.~Guo, D.~Gong, and Q.~Chen, ``Dynamic ensemble selection for
  imbalanced data streams with concept drift,'' \emph{IEEE transactions on
  neural networks and learning systems}, vol.~35, no.~1, pp. 1278--1291, 2022.

\bibitem{lu2019adaptive}
Y.~Lu, Y.-M. Cheung, and Y.~Y. Tang, ``Adaptive chunk-based dynamic weighted
  majority for imbalanced data streams with concept drift,'' \emph{IEEE
  Transactions on Neural Networks and Learning Systems}, vol.~31, no.~8, pp.
  2764--2778, 2019.

\bibitem{hu2025performance}
S.~Hu, Z.~Liu, and X.~He, ``Performance-bounded online ensemble learning method
  based on multi-armed bandits and its applications in real-time safety
  assessment,'' \emph{arXiv preprint arXiv:2503.15581}, 2025.

\bibitem{pao1994learning}
Y.-H. Pao, G.-H. Park, and D.~J. Sobajic, ``Learning and generalization
  characteristics of the random vector functional-link net,''
  \emph{Neurocomputing}, vol.~6, no.~2, pp. 163--180, 1994.

\bibitem{malik2023random}
A.~K. Malik, R.~Gao, M.~Ganaie, M.~Tanveer, and P.~N. Suganthan, ``Random
  vector functional link network: Recent developments, applications, and future
  directions,'' \emph{Applied Soft Computing}, vol. 143, p. 110377, 2023.

\bibitem{zhang2016comprehensive}
L.~Zhang and P.~N. Suganthan, ``A comprehensive evaluation of random vector
  functional link networks,'' \emph{Information sciences}, vol. 367, pp.
  1094--1105, 2016.

\bibitem{hager1989updating}
W.~W. Hager, ``Updating the inverse of a matrix,'' \emph{SIAM review}, vol.~31,
  no.~2, pp. 221--239, 1989.

\bibitem{bounebirat2017several}
F.~Bounebirat, D.~Laissaoui, and M.~Rahmani, ``Several explicit formulae of
  sums and hyper-sums of powers of integers,'' \emph{arXiv preprint
  arXiv:1712.07208}, 2017.

\bibitem{si2019powers}
D.~T. Si, ``The powers sums, bernoulli numbers, bernoulli polynomials
  rethinked,'' \emph{Applied mathematics}, vol.~10, pp. 100--112, 2019.

\bibitem{page1954continuous}
E.~S. Page, ``Continuous inspection schemes,'' \emph{Biometrika}, vol.~41, no.
  1/2, pp. 100--115, 1954.

\end{thebibliography}
\end{document}